\documentclass{article}

\usepackage{arabtex}

\usepackage{PRIMEarxiv}

\usepackage[utf8]{inputenc} 
\usepackage[T1]{fontenc}    
\usepackage{hyperref}       
\usepackage{url}            
\usepackage{booktabs}       
\usepackage{amsfonts}       
\usepackage{nicefrac}       
\usepackage{microtype}      
\usepackage{lipsum}
\usepackage{fancyhdr}       
\usepackage{graphicx}       
\graphicspath{{media/}}     

\usepackage[figuresright]{rotating}
\usepackage{times}
\usepackage{latexsym}
\usepackage{multirow}
\usepackage{lineno}
\usepackage{tabularx}
\usepackage{array,booktabs}
\usepackage{etoolbox}
\usepackage{microtype}
\usepackage{utf8}
\usepackage{graphicx}
\usepackage{verbatim}
\usepackage{utf8}
\usepackage{pdfpages}
\usepackage{float}
\usepackage[cmex10]{amsmath}
\usepackage{amssymb}
\usepackage{epstopdf}
\usepackage{amsthm,caption}

\renewenvironment{abstract}%
         {\centerline{\large\bf Abstract}%
          \begin{list}{}%
             {\setlength{\rightmargin}{0.6cm}%
              \setlength{\leftmargin}{0.6cm}}%
           \item[]\ignorespaces}%
         {\unskip\end{list}}

\title{AraCOVID19-SSD:  Arabic COVID-19 Sentiment and Sarcasm Detection Dataset
}

\author{
  Mohamed Seghir Hadj Ameur, Hassina Aliane \\
  Research and Development in Digital Humanities Division \\
  Research Centre on Scientific and Technical Information (CERIST) \\
  Algiers, Algeria\\
  \texttt{\{mhadjameur, ahassina\}@cerist.dz} \\
}

\begin{document}
\setcode{utf8}
\maketitle

\begin{abstract}
Coronavirus disease (COVID-19) is an infectious respiratory disease that was first discovered in late December 2019, in Wuhan, China, and then spread worldwide causing a lot of panic and death. Users of social networking sites such as Facebook and Twitter have been focused on reading, publishing, and sharing novelties, tweets, and articles regarding the newly emerging pandemic. 
A lot of these users often employ sarcasm to convey their intended meaning in a humorous, funny, and indirect way making it hard for computer-based applications to automatically understand and identify their goal and the harm level that they can inflect. Motivated by the emerging need for annotated datasets that tackle these kinds of problems in the context of COVID-19, this paper builds and releases AraCOVID19-SSD\footnote{\url{https://github.com/MohamedHadjAmeur/AraCovid19-SSD}} a manually annotated Arabic COVID-19 sarcasm and sentiment detection dataset containing 5,162 tweets. To confirm the practical utility of the built dataset, it has been carefully analyzed and tested using several classification models. 
\end{abstract}


\section{Introduction}
COVID-19 is a highly infectious respiratory \cite{lai2020severe} that was first identified in Wuhan, China, in late December 2019, and then declared as a global pandemic on March 2020 by the World Health Organization (WHO) \cite{di2020coronavirus}. Since its appearance governments around the world have adopted several protection measures such as closing borders, travel restrictions, quarantine, and containment. As of late July 2021, COVID-19 has caused more than 170 million confirmed cases and 3 million deaths worldwide\footnote{\url{https://www.worldometers.info/coronavirus/}}. Governments around the world have taken some urgent measures to stop the spread of the virus such as closing borders, self-isolation, quarantine, and social distancing. 

The severity of these measures along with the increased number of cases and deaths have significantly impacted people's morals causing a lot of uncertainty, grief, fear, stress, mood disturbances, and mental health issues \cite{10.1093pubmedfdaa106}. Many people relied on social networking sites such as Facebook and Twitter to express their feelings, thoughts, and opinions by publishing and sharing content related to this new emerging pandemic. The content that they shared often employed sarcasm  \footnote{Sarcasm can be defined as ‘a cutting, often ironic remark intended to express contempt or ridicule’ \url{www.thefreedictionary.com}} to convey their intended meaning in a humorous, funny, and indirect way making it hard for the computer-based applications to automatically understand and identify their goal and the harm level that they can cause. The presence of sarcastic phrases makes the task of sentiment analysis more difficult as the intended meaning is conveyed via indirect often humorous ways. This led the research community to devote a lot of interest and attention to the task of automatic sarcasm and sentiment detection.
As part of the efforts that are being made to create and share COVID-19 related datasets and tools \cite{Gautam_2006.11343, Elhadad_10.1007978-3-030-57796-4_25, ArabicTwitterDataset_2004.04315,  HADJAMEUR2021232}, this paper builds and releases a manually annotated Arabic COVID-19 sarcasm and sentiment detection dataset containing 5,162 tweets. The built dataset is carefully analyzed and tested using several classification models. 

The main contributions of this paper can be summarized as follows:
\begin{itemize}
    \item We collected, treated, and made available a large annotated Arabic COVID-19 Twitter sentiment and sarcasm detection dataset which can be very helpful to the research community.
    
    \item To the best of our knowledge, this is the first paper that shares an annotated dataset for both Arabic sentiment and sarcasm detection in the context of the COVID-19 pandemic. 

    \item We compared the results of multiple bag-of-words and pre-trained transformer baselines for the two considered tasks (sentiment analysis and sarcasm detection) and reported the obtained results.
\end{itemize}

The remainder of this paper is organized as follows: Section \ref{Related_Work} presents the sentiment and sarcasm detection research studies that have been published in the context of the COVID-19 pandemic. The details of our dataset collection, construction, and statistics are then provided in Section \ref{Dataset_section}. Then, in Section \ref{Exp_setup_section}, we present and discuss the tests we have done and the results we have obtained. Finally, In Section \ref{Conclusion}, we conclude our work and highlight some possible future works.

\section{Related Work}
\label{Related_Work}
In the last decade, the research studies that have been made in regards to Arabic sentiment analysis and sarcasm detection have increased significantly. As such, a large number of datasets have been built and shared to be used by the research community. In which concerns the major research studies that attempted to create Arabic sentiment analysis datasets;  
Rushdi et al. \cite{rushdi2011oca} presented an Arabic opinion mining dataset containing 500 movie reviews gathered from several blogs and web pages. Their dataset contains the same number of positive and negative instances, 250 each. The authors used several machine learning algorithms so as to provide baseline results for their annotated dataset.
Nabil et al. \cite{nabil-etal-2015-astd} presented the Arabic Social Sentiment Analysis Dataset (ASTD). It contains 10,000 Arabic tweets manually annotated with four labels: “objective”, “subjective positive”, “subjective negative”, “subjective mixed”. Their paper also presented the statistics of their constructed dataset as well as its baseline results.
Al-Twairesh et al. \cite{al2016arasenti} created two Arabic sentiment lexicons using a large tweets dataset containing $2.2$ million tweets. Their lexicons were generated using two methods and evaluated by using internal and external datasets. 
Aly et al. \cite{aly-atiya-2013-labr} created an Arabic sentiment analysis dataset containing over 63,000 book reviews, each review is rated on a scale of 1 to 5 stars. They provided baseline results for their dataset by testing it on the tasks of sentiment polarity and rating classification. 
Abu Kwaik et al. \cite{abu-kwaik-etal-2020-arabic} presented an Arabic sentiment analysis dataset containing 36,000 annotated tweets. The authors employed distant supervision and self-training approaches to annotate the collected tweets. They also released 8,000 tweets that have been manually annotated as a gold standard. 

For the task of sarcasm detection, several datasets have been published. 
Karoui et al. \cite{KAROUI2017161} created a sarcasm and irony dataset using political tweets from Twitter. They gather the tweets using politician names as keywords and classify them into ironic and non-ironic tweets. Their created dataset contains a total of 5,479 tweets, 1,733 of which are ironic and the remaining are non-ironic.
Ghanem et al. \cite{10.1145/3368567.3368585} created a shared task for Arabic irony detection consisting of binary classification of tweets as ironic or non-ironic. They released a dataset composed of 5,030  tweets regarding the Middle East and Maghreb regions' political events. Their tweets were composed of Modern Standard Arabic (MSA) as well as different Arabic regional dialects. 
Abbes et al. \cite{abbes-etal-2020-daict} created an irony-detection corpus (DAICT) that includes a  total of 5,358 annotated MSA and dialectal Arabic tweets. The tweets were collected on the basis of different hashtags regarding irony and sarcasm. Their classification included 3 labels: “Ironic”, “Not Ironic”, and “Ambiguous”.
Farha et al. \cite{farha2020arabic} presented “ArSarcasm”, an Arabic sarcasm detection dataset built by re-annotating an existing sentiment analysis dataset. Their dataset contains 10,547 tweets, $16\%$ of which are sarcastic. They used different baselines to test the utility of their dataset and reported the obtained results.

To the best of our knowledge there are no research studies that have attempted to build a sentiment analysis and sarcasm detection dataset that is devoted to the COVID-19 pandemic, thus, we believe that our dataset will be an important addition to the efforts that are being held to make more COVID-19 related datasets available for the research community.   

\section{Dataset}
\label{Dataset_section}
This section first presents the “AraCOVID19-SSD”\footnote{\url{https://github.com/MohamedHadjAmeur/AraCovid19-SSD}} dataset, its design goals, and the different classes that it contains. Then, it explains the process of tweets collection and annotation that has been adopted and provides the dataset’s statistics.

\subsection{“AraCOVID19-SSD” Dataset Description}
\label{Dataset_Description}
The “AraCOVID19-SSD” considers two tasks: sarcasm detection and sentiment analysis. The tasks' descriptions and their annotation details are provided in Table \ref{tab:classification_tweets}.

\begin{table*}[h]
	\small
	\centering
	\caption{“AraCOVID19-SSD” tasks, values, and their signification}
	{\def\arraystretch{2.5}\tabcolsep=2.5pt	
		\begin{tabular}{>{\centering\arraybackslash} m{25mm} >{\centering\arraybackslash}m{22mm} >{\centering\arraybackslash} m{95mm}}\hline
			\textbf{Tasks} & \textbf{Values} & \textbf{Explanation} \\\hline
			\textbf{Sarcastic} & Yes, No & Whether the tweet is sarcastic or not. \\\hline
			\textbf{Sentiment} & Positive, Negative, Neutral & Whether the tweet's Sentiment is Positive, Negative, or Neutral. \\\hline
		\end{tabular}
	}
	\label{tab:classification_tweets}
\end{table*}	

\begin{figure}[h]
	\centering
	\captionof{table}[foo]{Example of some Arabic tweets along with their respective annotations}
	\includegraphics[width=0.95\linewidth]{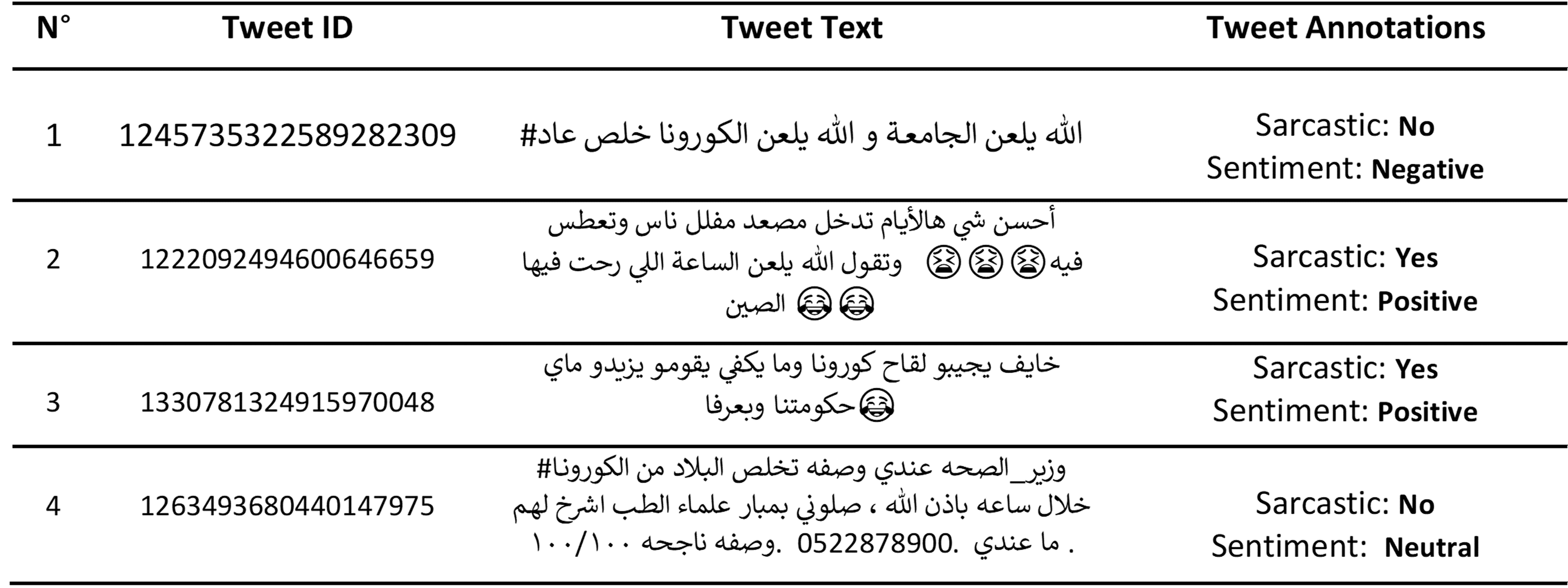}
	\label{img:exemple_tweets}
\end{figure}

All the 5,162 Arabic tweets of the “AraCOVID19-SSD” dataset are annotated for the two aforementioned tasks (Table \ref{tab:classification_tweets}). A small portion of the  “AraCOVID19-SSD” annotated tweets are illustrated in Table \ref{img:exemple_tweets}.

\subsection{Data Collection}
The first step that we followed to build the dataset was to prepare a set of keywords, then we retrieved the tweets based on those keywords. The keywords that we used were made to retrieve the largest possible number of tweets related to COVID-19, a portion of the keywords that we used are 				\<  فيروس، الفيروس،  الوباء، وباء،   كورونا، كوفيد، الكورونا، الكوفيد >\\

The retrieved tweets were filtered in the following way:
\begin{itemize}
	\item All the retweets of a given tweet were removed.
	\item Identical tweets that share the same textual content (when ignoring the tweets' links) were removed. This is done to ensure that the text of each considered tweet is unique.
	\item Very short tweets that contain less than $5$ Arabic words were filtered.
	\item Tweets were gathered within the period spanning from December 15, 2019, and December 15, 2020.
\end{itemize}
After the filtering step, we have ended up with a total of 300k unique Arabic tweets related to the COVID-19 pandemic.

\subsection{Data Annotation}
Due to the high cost of the annotation task, we only required each tweet to be annotated by one expert annotator. This allowed us to annotate a total of $5,162$ Arabic tweets from the 300k gathered tweets. We plan to continue annotating the remaining gathered tweets gradually according to our financial capacities.
The manual annotation task was carried out by providing the annotator with the full text of the tweet, including the links, and ask him/her to read the tweet, check the tweet's links if necessary, and annotate it for each one of the 2 labels (tasks). This results in a dataset in which each tweet is labeled for each one of the 2 tasks (as shown in Table \ref{img:exemple_tweets}).

\subsection{“AraCOVID19-SSD” Statistics}
The statistics of our “AraCOVID19-SSD” dataset are provided in Table \ref{table:stats}.

\begin{table*}[h]
	\small
	\centering
	\caption{Statistics about the number of annotated tweets for each task}
	{\def\arraystretch{2.5}\tabcolsep=2.5pt	
		\begin{tabular}{|c|c|c|}
            \hline
            \textbf{Tasks}                                 & \textbf{Annotation statistics} & \textbf{Total Tweets}                                                                      \\ \hline
            \textbf{Sarcastic}                        & \textbf{Yes}: 1802  \space\space\space    \textbf{No}: 3360 & 5162 \\
            \textbf{Sentiment}                    & \textbf{Positive}: 1964  \space\space\space   \textbf{Negative}: 1197  \space\space\space   \textbf{Neutral}: 2001             & 5162                    \\
            \hline                                                                   
        \end{tabular}
        }
        \label{table:stats}
        \end{table*}

As shown in Table \ref{table:stats}, the two considered tasks contain more than $1000$ instances for each one of their values, which helps train robust classification models.

\section{Tests and Results}
\label{Exp_setup_section}
Our tests aim at evaluating the quality of our annotated dataset and provide baseline results for the two tasks that it includes. To this end, for both sarcasm detection and sentiment analysis tasks, several deep learning and bag-of-words models were trained and tested. In the following, first,  we will present the Arabic preprocessing that we performed, and the different models that we considered. Then, we will report and discuss the results of our performed tests.

\subsection{Preprocessing}
We applied a basic preprocessing to all the collected Arabic tweets, which includes:
\begin{itemize}
	\item The removal of diacritical marks.
	\item The removal of elongated and repeated characters.
	\item Arabic characters normalization.
	\item  The removal of links and users' references (users' notifications). 
	\item Tweets tokenization in which punctuation, words, and numbers are separated. 
\end{itemize}
We note that this preprocessing has been used only when training the baseline models (Section \ref{Considered_Models}); it has not been used for the annotation task nor in the final dataset.

\subsection{Considered Models}
\label{Considered_Models}
Aside from the classical bag-of-words models, pretrained transformer models have been recently used in many NLP tasks and have continuously achieved new state-of-the-art results \cite{DBLP:BERT_abs-1810-04805}. In the following, we will highlight both the bag-of-words and the transformer models that we considered in our tests.

\subsubsection{Transformer Models}
In our experiments we used three pretrained transformer models:
\begin{itemize}
	\item AraBERT\footnote{\url{https://huggingface.co/aubmindlab/bert-base-arabertv02}}: A  BERT (Bidirectional Encoder Representations from Transformers) model \cite{DBLP:BERT_abs-1810-04805} pretrained on $200$ million Arabic MSA sentences gathered from different sources \cite{antoun2020arabert}. 
	\item Multilingual BERT (mBERT)\footnote{\url{https://huggingface.co/bert-base-multilingual-cased}}: A BERT-based model \cite{DBLP:BERT_abs-1810-04805} pretrained on the first $104$ major Wikipedia languages\footnote{\url{https://meta.wikimedia.org/wiki/List_of_Wikipedias}}. 
	\item XLM-Roberta \footnote{\url{https://huggingface.co/transformers/model_doc/xlmroberta.html}}: A large multi-lingual language model, trained on $2.5$TB of filtered Common Crawl data \cite{conneau-etal-2020-unsupervised}.
\end{itemize}

\subsubsection{Bag-of-Words Models}
In our experiments we considered three bag-of-words models:
\begin{itemize}
	\item Support Vector Machines (SVMs) \cite{hearst1998support}: are discriminative classifiers that use maximum-margin hyperplanes (support vectors) to classify high-dimensional data into a set of predefined categories.
	
	\item Random Forests model \cite{Breiman2001}: 
	is an extension to the standard decision tree \cite{myles2004introduction} introduced to tackle the over-fitting problem that usually occurs when a decision tree learns highly irregular patterns as a consequence of growing too deep. It constructs multiple trees from random sub-samples of the same training data. Then, the final prediction is made by averaging the predictions of all the trained trees.

	\item Logistic Regression \cite{wright1995logistic}: is a process of modeling the probability of an outcome given an input variable. It is useful for classification problems, where the goal is to determine if a new instance fits best into a given category. 
\end{itemize}	

\subsection{Software and Tools}
The implementation of the different models have been done using the following libraries:
\begin{itemize}
	\item Scikit-learn  \cite{scikit-learn}\footnote{\url{https://scikit-learn.org/stable/}} is a python-based  machine learning library. We used it to train the bag-of-words baselines and to evaluate the performance of all the considered models.
	
	\item Flair \cite{Flair_akbik2018coling}\footnote{\url{https://github.com/flairNLP/flair}}: is a framework for building state-of-the-art NLP models. We used it to train our classification models.
	
	\item Huggingface-transformers \cite{wolf-etal-2020-transformers}\footnote{\url{https://github.com/huggingface/transformers}}: is a framework for building and pretraining different state-of-the-art NLP models. We used it to test our pretrained models.
	
	\item PyTorch \cite{PyTorch_NEURIPS2019_9015}\footnote{\url{https://github.com/pytorch/pytorch}} is an open-source library designed for implementing deep neural networks. We used it as a backend for both the Huggingface-transformers and the Flair frameworks.
	
\end{itemize}

\subsection{Evaluation Methodology}
To evaluate the performance of our considered classification models, we have used a stratified 5-fold cross-validation method. This is done by randomly partitioning the instances of each one of our dataset's tasks into 5 disjoint sets of equal size. In this five-fold cross-validation, five experiments are performed, in each one, one of the five sets is selected for testing, and the remaining four are used for training. For each experiment, the weighted F-score is calculated, and finally, the average F-score for all the five experiments (the 5-folds) is reported.

\subsection{Results and Discussion}
\label{Test_section}
The results of the experiments that we have performed in regards to the tasks of sarcasm detection and sentiment analysis are provided in Table \ref{table:results}.

\begin{table*}[h]
	\footnotesize
	\centering
	\caption{Weighted F-score results for the considered classification models}
	{\def\arraystretch{2.5}\tabcolsep=2.5pt	
		\begin{tabular}{|c|c|ccc|ccc|}
			\cline{3-8}
			 \multicolumn{1}{c}{} & \multicolumn{1}{c|}{} & \multicolumn{3}{c|}{\textbf{Sarcasm}} & \multicolumn{3}{c|}{\textbf{Sentiment}} \\\cline{3-8}
			 
			 \multicolumn{1}{c}{}&\multicolumn{1}{c|}{} & Recall & Precision & F-score &  Recall & Precision & F-score \\\cline{1-8} 
			 
			 \multirow{3}{*}{Bag-of-words Models} 
			 & \textbf{Logistic Regression} 	 &  0.9538  & 0.9540  & 0.9538 & 0.9069 & 0.91 & 0.9072 \\
			 & \textbf{Random Forest} 		 &  0.9511  & 0.9512  & 0.9511 & 0.8994 & 0.9019 & 0.8997 \\
			 & \textbf{SVM} 					 &  0.9597  & 0.9599  & \textbf{0.9597} & 0.9223 & 0.9243 & 0.9224 \\\cline{1-8} 
			 
			 \multirow{3}{*}{Transformers Models} 
			 & \textbf{XLM-Roberta} 			& 0.9361 & 0.9369 & 0.9360 &  0.9186  & 0.9211  & 0.9187 \\
			  & \textbf{BERT-multi} 			& 0.9416 & 0.9431 & 0.94162 &  0.8928  & 0.89592  & 0.8930 \\
			  & \textbf{Arabert}  			& 0.9527 & 0.9535 & 0.9527 &  0.9225  & 0.9234  & \textbf{0.9226} \\\cline{2-8} 
			 \cline{1-8}                              
		\end{tabular}
		}
		\label{table:results}
	\end{table*}

The experiments that we have performed show that high-level classification results are achieved for both the sentiment and sarcasm detection tasks. Indeed, all the tested models surpassed $0.89$ f-score, we believe that the high f-scores that have been achieved are mainly due to the richness of the dataset (the high number of instances in each class of the considered tasks).   
We can also observe that the SVM model and the Arabert transformer model gave the best performance by reaching an f-score of more than $0.95$ on the sarcasm detection task and more than $0.92$ on the sentiment analysis task.

The quality of the obtained results reflects the importance of having a large annotated dataset and confirms our adopted annotation schema's practical utility.

\section{Conclusion}
\label{Conclusion}
In this paper, we have presented and published “AraCOVID19-SSD” an Arabic COVID-19 sentiment analysis and sarcasm detection dataset. The dataset contains 5,162 Arabic tweets; each tweet is annotated for two tasks: sentiment analysis and sarcasm detection.  All the dataset's tweets have been manually annotated and validated by human annotators. The quality of the final annotated dataset has been examined via several bag-of-words and transformer models. The considered models were trained and tested using the developed dataset and the obtained results were reported.
As future work, we plan to continue enriching the annotated dataset with new tweets to keep it up-to-date with the latest events and discussions that are being shared on Twitter in regards to the COVID-19 pandemic.

\bibliographystyle{unsrt}

\end{document}